\documentclass[letterpaper, 10 pt, conference]{ieeeconf}
\IEEEoverridecommandlockouts
\usepackage{cite}
\usepackage{amsmath,amssymb,amsfonts}
\usepackage{graphicx}
\usepackage{booktabs}
\usepackage{hyperref}
\usepackage{xcolor}
\usepackage{tikz}
\usetikzlibrary{arrows.meta,positioning,fit,calc}

\title{\LARGE \bf
Compositional Shift Algebra:\\
Extrapolating Mixed Robot Shifts Without Mixed Finetuning
}

\author{Jinting Hang$^{\ast}$ and Zhenhui Cai%
\thanks{$^{\ast}$Corresponding author:
\texttt{jinting.hang@harvest-praxis.com}.
Z.~Cai: \texttt{philip.cai@harvest-praxis.com}.}%
\\
\normalsize Harvest Praxis%
}

\begin{document}
\maketitle
\thispagestyle{empty}
\pagestyle{empty}

\begin{abstract}
Robot deployments rarely change one mechanism at a time: cameras, action interfaces, and physical dynamics often shift together.
Prior adaptation recipes either finetune a new model for every mix or attempt to select which module to update.
We instead learn shift operators on a modular stack $z{=}E(o)$, $a{=}g(z,u)$, $z'{=}f(z,a)$ and compose them.
Compositional Shift Algebra (CSA) fits single-factor observation, policy, and dynamics operators from exact-reset probes, then extrapolates held-out mixed shifts by operator composition---without mixed-shift finetuning.
On ManiSkill StackCube, residual CSA matches an oracle mixed inverse on held-out mixes (success $1.0$ over $10$ seeds) while beating best-single / zero-shot / parameter-average baselines by ${\approx}67$\,pp.
RGB-D vision-in-the-loop composition remains near oracle and far above non-compositional arms; a delay commutator stress shows ordered necessity for policy$\times$delay.
On a second task (PickCube), residual CSA again reaches compose $1.0$ vs.\ $0.33$ non-compositional ($n{=}10$), and an L1 vision controller without privileged cube/goal poses or grasp flags in the control loop retains compose $0.95$ vs.\ $0.00$.
Main-track upgrades freeze PushCube ($+33$\,pp), PegInsertion joint8 / pose7 EE ($+67$\,pp each), and thin BC under frozen CSA ($+67$\,pp); deeper BC and fair adapt baselines still need compose ($+67$\,pp each), vision-localized BC needs compose ($+56$\,pp), and delay favors ordered/few-shot deploy.
We report Intervention-Gated Adaptation as a negative control.
\end{abstract}

\section{INTRODUCTION}

A new camera changes pixels; a new teleoperator changes commanded actions; a new payload or action delay changes transitions.
Real deployments combine these factors.
Finetuning a separate model for every mix does not scale, and freezing a fixed module set fails when the wrong mechanism changed.

A previous line asked which module to update (Intervention-Gated Adaptation, IGA).
On StackCube, learned gating did not clear a method-level Go/No-Go bar once oracle-sparse failed to beat strong fixed arms (Sec.~\ref{sec:iga}).
The productive question is therefore not only \emph{what} to update, but whether mixed shifts are algebraically recoverable from single-factor operators.

We study Compositional Shift Algebra (CSA).
A deployment shift is a mechanism-indexed operator $\Phi_\sigma{=}(\Phi^E_\alpha,\Phi^g_\beta,\Phi^f_\gamma)$ acting on modules.
The algebraic hypothesis is
\begin{equation}
  \Phi_{\sigma\circ\tau} \approx \Phi_\sigma \circ \Phi_\tau
  \label{eq:compose}
\end{equation}
on a declared support.
Training may observe singles and at most a sparse consistency mix; evaluation must include held-out pairwise/triple compositions never used as mixed training targets.

\begin{figure}[t]
  \centering
  \resizebox{0.98\columnwidth}{!}{%
  \begin{tikzpicture}[
    font=\scriptsize,
    >=Stealth,
    box/.style={draw=black!60, rounded corners=1pt, align=center,
      inner sep=2pt, minimum height=0.85cm},
    lab/.style={font=\tiny\bfseries}
  ]
    \node[box, fill=black!4, text width=2.0cm] (A)
      {A. Single-factor fits\\$\Phi_E,\Phi_g,\Phi_f$\\obs / policy / dyn};
    \node[box, fill=cyan!8, text width=2.1cm, right=0.25cm of A] (B)
      {B. Shift algebra\\$\Phi_{\sigma\circ\tau}$\\compose generators};
    \node[box, fill=orange!8, text width=2.1cm, right=0.25cm of B] (C)
      {C. Held-out mixes\\obs$\circ$dyn / pol$\circ$dyn\\triple};
    \node[box, fill=green!8, text width=2.2cm, right=0.25cm of C] (D)
      {D. Closed loop\\compose $\approx$ oracle\\$>$ best single};
    \draw[->, thick] (A) -- (B);
    \draw[->, thick] (B) -- (C);
    \draw[->, thick] (C) -- (D);
  \end{tikzpicture}}
  \caption{CSA fits single-factor shift operators, composes them for held-out mixes, and evaluates closed-loop success without mixed finetuning.}
  \label{fig:teaser}
\end{figure}
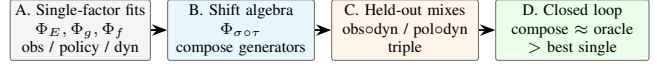

\paragraph{Contributions.}
\begin{enumerate}
  \item We formulate CSA for modular robot stacks and an additive-generator MVP with an explicit delay commutator stress.
  \item Held-out StackCube mixes are closed-loop corrected by composing single-factor residual operators, matching oracle and beating non-compositional baselines under multi-seed evaluation; RGB-D vision and delay order stresses extend the claim.
  \item On PickCube, residual CSA transfers under a shared ShiftSpec, and L1 vision control removes privileged cube/goal/contact from the loop while retaining compositional gains (with disclosed workspace-camera goal fallback; P1 fusion removes that default).
  \item Main-track upgrades: PushCube and Peg (joint8 + pose7 EE) extend the task family; thin/deep BC under frozen CSA and fair adapt baselines upgrade the controller claim; ordered compose for delay; IGA is a negative result.
\end{enumerate}

\section{PROBLEM FORMULATION}

We use the modular structural model
\begin{align}
  z_t &= E(o_t), \label{eq:e}\\
  a_t &= g(z_t, u_t), \label{eq:g}\\
  z_{t+1} &= f(z_t, a_t), \label{eq:f}
\end{align}
with optional rendering $o_t{=}r(z_t,c_t)$.
A binary mask $m\in\{0,1\}^3$ indexes observation, policy, and dynamics interventions.
Unlike IGA, the object of learning is not $m$ alone but operators $\Phi$ that transform module parameters (or closed-loop corrections) so that~\eqref{eq:compose} recovers unseen mixes.

\paragraph{Factor-family holdout.}
Seed holdout alone is invalid.
We train on singles $(1,0,0)$, $(0,1,0)$, $(0,0,1)$ and optionally obs$\circ$policy for consistency; we forever hold out obs$\circ$dyn, policy$\circ$dyn, and the triple for mixed finetuning.

\paragraph{Go / No-Go.}
Composition must stay within $5$\,pp of an oracle mixed inverse on held-out pairs, beat the best non-compositional baseline by ${\ge}8$\,pp, and pass a commutator stress (commutative pairs near-insensitive; delay-like pairs show ordered necessity).

\section{THREAT MODEL AND SCOPE}
\label{sec:threat}

CSA is not a claim about arbitrary open-world shift.
We restrict the adversary (and the benchmark) as follows.

\paragraph{In scope.}
Factorized shifts drawn from a known mechanism family: post-render observation maps (pixel translation / appearance), invertible policy remaps (scale/bias/linear), and low-order dynamics edits (mass, friction, damping) and/or constant action delay.
The deployment may apply any \emph{subset} of these factors, including held-out mixes never seen as joint training targets.
Exact-reset probes make single-factor identification well-posed; closed-loop success is the primary utility.

\paragraph{Out of scope (explicit non-goals).}
Unmodeled contact modes, unannounced new object categories, nonstationary delay, learned end-to-end policies as the primary controller, and Wan/DiT-internal ``physics'' factors.
If a factor is non-commutative with others (action delay), Abelian additive composition is \emph{not} presumed; ordered deploy is required (Sec.~\ref{sec:delay}).

\paragraph{What ``composition works'' means.}
On held-out mixes, a composed correction must (i) approach an oracle that knows the mixed inverse and (ii) beat non-compositional arms that see the same singles but do not compose.
A gap to oracle without a gap to non-compositional arms is insufficient for an algebra claim.

\paragraph{Honest failure modes we already hit.}
(1) Policy$\times$delay($=2$): Abelian residual compose regresses; ordered lead recovers oracle.
(2) PickCube residual/L1: closed-loop dynamics mix is often $0$---compose wins via observation/policy correction, not action residual blending.
(3) L1 agent-only / shoulder-only goal: wrist occlusion or empty base view collapses success; P1 hand$\to$base fusion removes the third-person workspace cam while matching workspace compose (Sec.~\ref{sec:pickcube}).

\section{METHOD}

\subsection{Additive generators (MVP)}
For flattened mechanism parameters $M$, CSA uses
\begin{equation}
  \Phi_\theta(M) = M + \sum_i \theta_i G_i,
\end{equation}
with coordinate addition as Abelian composition.
This approximation is intentional: non-commutative factors (action delay) are tested separately rather than forced into addition.

\subsection{Single-factor identification}
On ManiSkill tasks with exact state restore:
\begin{itemize}
  \item \textbf{Observation:} fit an inverse of the post-render pixel translation from paired RGB.
  \item \textbf{Policy:} fit an inverse of the action remapping from commanded vs.\ executed probes.
  \item \textbf{Dynamics:} fit an affine one-step residual under source vs.\ dynamics-shifted rollouts; $G{=}M_{\mathrm{shift}}{-}M_{\mathrm{source}}$. Closed-loop correction solves for an action whose shifted predictor matches the source target; $(\mathrm{ridge},\mathrm{mix})$ are selected on held-out match.
\end{itemize}

\subsection{Composition at test time}
\label{sec:compose-algo}
Given estimated coordinates $(\hat\alpha,\hat\beta,\hat\gamma)$ (oracle mask or few-shot probes), CSA builds a correction state
\begin{equation}
  \widehat\Phi
  =
  \Phi^E_{\hat\alpha}\,
  \circ\,
  \Phi^g_{\hat\beta}\,
  \circ\,
  \Phi^f_{\hat\gamma}
\end{equation}
(or the ordered variant required by a commutator stress) and deploys it without mixed finetuning.
When the dynamics residual mix weight selected by closed-loop validation is near zero, $\Phi^f$ contributes negligibly and composition reduces to observation/policy inversion---still a valid instance of~\eqref{eq:compose} on the active factors, but not evidence that residual dynamics mixing drove the gain.
Arms: compose, compose fewshot, best single, param average, oracle mixed, zero shot.

\subsection{Vision-in-the-loop and L1}
For observation-containing held-outs, object XYZ for the FSM comes from RGB-D / segmentation on obs-corrected depth.
End-effector XYZ remains proprioceptive; simulator success stays scoring-only.
\textbf{L1} further removes privileged cube pose, goal pose, and grasp flags from control: cube from agent RGB-D+seg; goal from a camera-visible marker (agent cam, else third-person workspace cam); grasp/place from gripper--geometry heuristics.
L1 therefore de-privileges \emph{state channels used by the controller}, not the entire evaluation stack.

\subsection{Delay commutator stress}
\label{sec:delay}
Declared-commutative obs$\circ$policy should be order-insensitive on the action stream when the dynamics residual is near identity.
Policy$\times$action-delay should change open-loop commands under dyn$\to$pol vs.\ pol$\to$dyn factorization order.
This is the operational test that Abelian addition is an MVP, not a universal algebra.
\section{EXPERIMENTAL SETUP}

\paragraph{Environment.}
ManiSkill StackCube, PickCube, PushCube, and PegInsertionSide (joint-space residual CSA), CPU PhysX, exact-reset isolation, factorized ShiftSpec interventions (post-render observation, action remapping, mass/friction/damping and/or action delay).

\paragraph{Protocols.}
Residual CSA multi-seed; RGB-D vision CSA multi-seed; delay commutator; PickCube residual / L1 / P1 fusion; PushCube residual / hard ShiftSpec; Peg joint8 MP-replay and Peg pose7 online EE convert; P3/P3-b/P3-c BC under frozen CSA; P3 adapt baselines; P3-d vision-localized BC.
Primary metrics: closed-loop task success and localization error under obs correction.

\paragraph{Baselines.}
Zero-shot identity; best active single; parameter average of singles; oracle mixed inverses (upper bound, not deployable under the claim).
Under frozen CSA we also compare source-only compose-BC to pooled train-singles BC and single-head output average (fair non-algebraic adapts).
Target-mix finetune is reported only as a disclosed upper bound: it \emph{uses} held-out mix labels and is not a fair CSA competitor.

\paragraph{Reporting.}
Headline numbers are frozen under \texttt{docs/CSA\_PICKCUBE\_L1\_FROZEN.md}, \texttt{docs/CSA\_MAINTRACK\_UPGRADE.md}, and the corresponding \texttt{results/.../main/} directories; we do not retune seeds to improve tables after freezing.
\section{RESULTS}

\subsection{Held-out residual composition (StackCube)}
\begin{figure}[t]
  \centering
  \includegraphics[width=0.92\columnwidth]{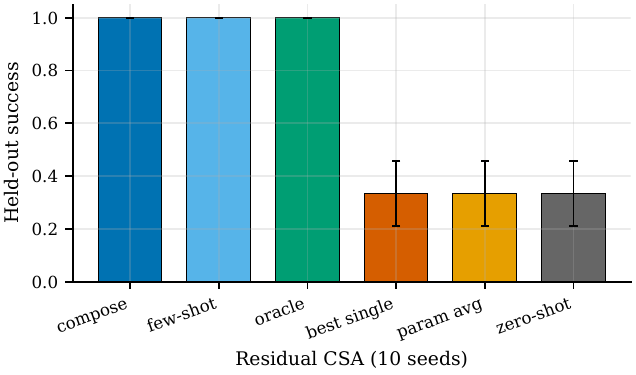}
  \caption{Residual CSA held-out success by arm (10 seeds; mean with 95\% CI).}
  \label{fig:residual}
\end{figure}

Figure~\ref{fig:residual} and Table~\ref{tab:residual} summarize $10$ independent fit+eval seeds under held-out mixes (seed-level Go rate $1.0$).
Compose and few-shot compose match oracle at success $1.0$, while best-single / param-average / zero-shot remain at $0.33$ ($+67$\,pp).
Residual next-state improvement ratio averages $0.35$ (95\% CI $[0.25,0.45]$).

\begin{table}[t]
\centering
\caption{Residual CSA multi-seed held-out closed-loop success (10 seeds; StackCube).}
\label{tab:residual}
\small
\begin{tabular}{lccc}
\toprule
Arm & Success & vs.\ oracle & vs.\ best non-comp. \\
\midrule
compose & $1.00$ & $0$\,pp & $+67$\,pp \\
compose fewshot & $1.00$ & $0$\,pp & $+67$\,pp \\
oracle mixed & $1.00$ & --- & --- \\
best single / param avg / zero-shot & $0.33$ & $-67$\,pp & --- \\
\bottomrule
\end{tabular}
\end{table}

\subsection{Vision-in-the-loop composition (StackCube)}
\begin{figure}[t]
  \centering
  \includegraphics[width=0.78\columnwidth]{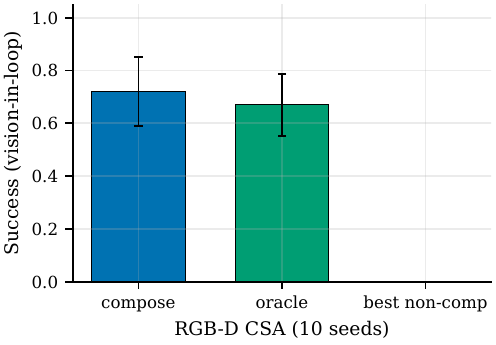}
  \caption{Vision-in-the-loop CSA success (10 seeds; mean $\pm$ 95\% CI).}
  \label{fig:vision}
\end{figure}

Under obs-containing held-outs with RGB-D cube localization ($10$ seeds $\times$ $5$ episodes), compose success is $0.72$ $[0.59,0.85]$ (Fig.~\ref{fig:vision}, Table~\ref{tab:vision}), matching or exceeding the vision+oracle-obs arm on average ($0.67$ $[0.55,0.79]$), with $+72$\,pp over the best non-compositional arm.
Per-seed Go rate is $0.6$; the aggregate claim remains GO.
Source depth localization mean cube error is ${\approx}2.0$\,cm.

\begin{table}[t]
\centering
\caption{Vision CSA (RGB-D) multi-seed summary on obs-containing held-outs.}
\label{tab:vision}
\small
\begin{tabular}{lc}
\toprule
Metric & Value \\
\midrule
compose success (mean, 95\% CI) & $0.72$ $[0.59,0.85]$ \\
oracle success & $0.67$ $[0.55,0.79]$ \\
gain over best non-compositional & $+72$\,pp \\
compose / zero-shot pose-error ratio & $0.33$ \\
per-seed GO rate & $0.6$ \\
decision / claim & GO / KEEP VISION CSA \\
\bottomrule
\end{tabular}
\end{table}

\subsection{Delay / commutator stress}
\begin{figure}[t]
  \centering
  \includegraphics[width=0.98\columnwidth]{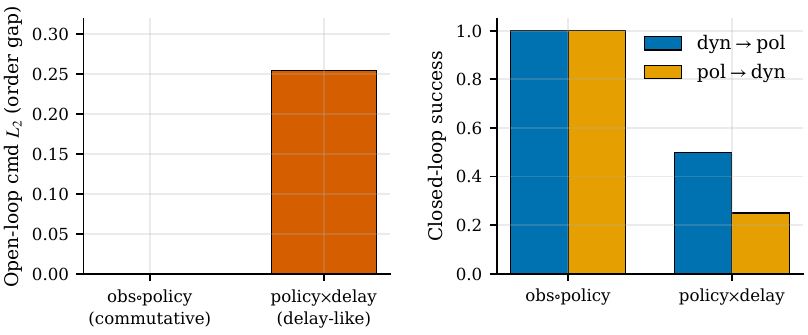}
  \caption{Delay commutator stress. Left: open-loop command divergence under factorization order. Right: closed-loop success for dyn$\to$pol vs.\ pol$\to$dyn.}
  \label{fig:delay}
\end{figure}

Additive algebra commutators are numerically $0$ (Abelian MVP).
Figure~\ref{fig:delay}: open-loop command $L_2$ between factorization orders is $0.00$ for obs$\circ$policy and $0.25$ for policy$\times$delay($=2$), while closed-loop compose prefers dyn$\to$pol ($0.50$ vs.\ $0.25$).

\paragraph{Harder held-out: policy$\times$delay($=2$).}
Abelian residual composition fails here (success $0.50$ below policy-only $0.83$).
An ordered (non-Abelian) deploy recovers $0.83$, matching oracle and beating the residual ablation by $+33$\,pp.

\subsection{Second task: PickCube residual and L1 vision}
\label{sec:pickcube}

\paragraph{Compositional shifts on a second task (PickCube).}
We reuse the same ShiftSpec factors (observation pixel shift, policy remapping, dynamics residual) on ManiSkill \texttt{PickCube-v1}.
Single-factor operators are fit only; held-out mixed masks are never finetuned.
With residual dynamics CSA over $n{=}10$ seeds, composed closed-loop success is $1.00$ versus $0.33$ for the best non-compositional arm (oracle $1.00$); the closed-loop residual mix is $0$ on all seeds, so the gain is carried by observation/policy composition rather than action residual blending.

\paragraph{L1: vision control without pose/contact privileges.}
We further remove privileged cube pose, goal pose, and grasp flags from the \emph{control} loop.
Cube XYZ comes from agent RGB-D + segmentation; the green goal marker is made camera-visible (ManiSkill otherwise hides it) and localized from segmentation, falling back to a third-person workspace camera when the wrist view is occluded; grasp/place use gripper--geometry heuristics.
Simulator success remains scoring-only; the end-effector is proprioceptive; the FSM is scripted.
Over $n{=}10$ seeds, composed success is $0.95$ $[0.88,1.02]$ versus $0.00$ for non-compositional arms (oracle $0.95$), with source localization error $\approx 2.4$\,cm (cube) / $2.2$\,cm (goal).

\paragraph{Ablations (short).}
Privileged contact flags do not improve composed success ($+0$\,pp).
Restricting goal localization to the agent camera drops success by $67$\,pp (localization-fail rate $0.67$), documenting the workspace-camera dependency.
Privileged goal XYZ matches L1 success on this set ($+0$\,pp).


\begin{table}[t]
\centering
\caption{PickCube CSA progression (held-out mixes; compose vs.\ best non-compositional). Numbers from \texttt{docs/CSA\_PICKCUBE\_L1\_FROZEN.md}.}
\label{tab:pick}
\small
\begin{tabular}{lcc}
\toprule
Protocol & Compose & Best non-comp \\
\midrule
Residual CSA ($n{=}10$) & $1.00$ & $0.33$ \\
Vision (goal privileged) & $1.00$ & $0.00$ \\
L1 vision ($n{=}10$) & $0.95$ $[0.88,1.02]$ & $0.00$ \\
\bottomrule
\end{tabular}
\end{table}

Table~\ref{tab:pick} summarizes the PickCube ladder.
Closed-loop residual mix is $0$ on all $10$ PickCube residual / L1 seeds: the compose win is carried by observation/policy composition, not action residual blending.
Short ablations (compose): privileged contact $+0$\,pp vs.\ L1; agent-only goal $-67$\,pp (loc-fail $0.67$); privileged goal $+0$\,pp.
P1 fusion on \texttt{panda\_wristcam} (hand$\to$base, no \texttt{render\_camera}): compose $1.00$ vs.\ wrist-only $0.67$ ($+33$\,pp) and shoulder-only $0.33$; matches the L1 workspace row at $1.00$ (loc-fail $0$).

\paragraph{Reading the PickCube result.}
PickCube strengthens the \emph{transfer} claim (shared ShiftSpec, second task) and the \emph{privilege} claim (L1), but it does not strengthen the claim that affine dynamics residuals are the active ingredient---because mix$=0$.
StackCube residual multi-seed, where mix is often selected near $1$, remains the primary dynamics-composition evidence.
Legacy L1 used a workspace-camera fallback under wrist occlusion; P1 shows robot-cam fusion recovers the same compose rate without that third-person default---still not a claim of learned detection or shoulder-only localization.

\subsection{Main-track upgrades (P1--P3, PushCube, Peg)}
\label{sec:upgrade}
Locked path P0$\to$P1$\to$(P2$|$P3)$\to$P4; numbers frozen under \texttt{docs/CSA\_MAINTRACK\_UPGRADE.md}.

\begin{table}[t]
\centering
\caption{Main-track upgrade closed-loop held-out success (compose vs.\ best non-compositional).}
\label{tab:upgrade}
\small
\begin{tabular}{lccc}
\toprule
Gate / protocol & Compose & Best non & $\Delta$ \\
\midrule
P1 L1 fusion (no workspace cam) & $1.00$ & $0.67$ & $+33$\,pp \\
P2 PushCube residual ($n{=}9$) & $1.00$ & $0.67$ & $+33$\,pp \\
P2-hard PushCube (delay$=2$) & $0.44$/$0.67^\dagger$ & $0.22$ & $+22$\,pp \\
P2 Peg joint8 MP-replay ($n{=}9$) & $1.00$ & $0.33$ & $+67$\,pp \\
P2 Peg pose7 EE online convert ($n{=}9$) & $1.00$ & $0.33$ & $+67$\,pp \\
P3 thin BC under frozen CSA & $1.00$ & $0.33$ & $+67$\,pp \\
P3-b thin BC, no phase scaffold & $0.67$ & $0.11$ & $+56$\,pp \\
P3-c deep BC under frozen CSA & $1.00$ & $0.33$ & $+67$\,pp \\
P3 adapt: compose vs.\ mixed-singles & $1.00$ & $0.33$ & $+67$\,pp \\
P3-d vision-localized BC ($n{=}9$) & $0.56$ & $0.00$ & $+56$\,pp \\
\bottomrule
\end{tabular}\\[0.15em]
{\scriptsize $^\dagger$Abelian compose / few-shot; delay favors ordered/few-shot (Sec.~\ref{sec:delay}).}
\end{table}

Table~\ref{tab:upgrade}: P1 removes the third-person workspace cam via hand$\to$base fusion.
P2 shows the algebra is not Stack/Pick-twin-specific (PushCube) and extends to orientation-heavy PegInsertion via \emph{8D joint} MP-trajectory replay and via \emph{7D} \texttt{pd\_ee\_delta\_pose} with online joint$\to$EE conversion (ManiSkill official convert; CSA hooks the student steps)---not a hand-rolled reactive pose servo.
P2-hard reconfirms non-Abelian delay stress on PushCube; method packaging prefers ordered lead / few-shot when delay$>0$ (\texttt{docs/CSA\_ORDERED\_COMPOSE.md}).
P3/P3-b/P3-c: source-only BC under \emph{frozen} CSA still needs compose (thin/deep $+67$\,pp; no-phase $+56$\,pp).
P3-d: at test time cube/goal come from RGB-D$+$segmentation (obs-corrected), not GT poses; compose $0.56$ vs.\ zero-shot/scratch $0.00$ ($+56$\,pp), below the FSM ceiling but still CSA-dependent.
Fair adapts (pooled train-singles; single-head average) stay ${\le}0.33$; disclosed target-mix finetune (uses held-out labels) also $0.33$ here---not a fair competitor.
P4 real-robot transfer remains hardware-blocked (\texttt{docs/CSA\_P4\_REAL.md}).

\section{NEGATIVE RESULT: INTERVENTION-GATED ADAPTATION}
\label{sec:iga}

IGA aimed to infer a multi-label changed-module mask and activate sparse adapters.
On exact-reset StackCube closed-loop smoke, sparse routing can preserve success when full updates interfere, but dynamics-module accuracy under the source-calibrated residual gate is only $0.71$, so the learned gate is NO-GO for a method claim.
We retain IGA as a diagnostic/negative control: \emph{knowing which module changed is not the same as having operators that compose on held-out mixes}.

\section{DISCUSSION}
\label{sec:discuss}

\paragraph{Why composition can beat module selection.}
Selection answers a discrete attribution question; CSA answers an extrapolation question.
When singles are identifiable and the support is approximately Abelian, held-out mixes are reachable by operator product without collecting mixed finetuning data.
When support is non-Abelian, selection still does not invent the missing ordered inverse---as delay shows.

\paragraph{Main-track upgrade path (declared).}
To move beyond a controlled-simulation algebra demo we prioritize, in order:
(P1) reduce workspace-camera dependence for L1 goals---\textbf{done}: hand$\to$base fusion GO;
(P2) third task / harder shifts---\textbf{done}: PushCube $+33$\,pp, Peg joint8 $+67$\,pp, Peg pose7 EE $+67$\,pp, PushCube-hard delay stress;
(P3) thin learned interface under frozen CSA---\textbf{done}: compose$+$thin $g$ reaches $1.00$ vs.\ $0.33$ zero-shot/scratch ($+67$\,pp); P3-b no-phase still $+56$\,pp; P3-c deep BC and fair adapt baselines still $+67$\,pp over zs/mixed-singles; P3-d vision-localized BC $+56$\,pp over zs/scratch;
(P4) small real-robot transfer when hardware is available (checklist: \texttt{docs/CSA\_P4\_REAL.md}).
We do not treat additional same-protocol seeds as an upgrade.
Sec.~\ref{sec:upgrade} freezes the upgrade ladder numbers, including parallel A/B/C protocols (deep BC, adapt baselines, ordered-compose packaging).

\section{RELATED WORK}

Domain randomization and multi-environment RL expand coverage but do not compose identified mechanism operators at test time~\cite{tobin2017domain}.
Parameter-efficient adaptation and mixture-of-experts select or blend modules~\cite{hu2022lora}; CSA instead extrapolates unseen mixes by algebraic composition of single-factor corrections.
Causal representation and intervention benchmarks motivate our factorized ShiftSpec generator and exact-reset isolation~\cite{scholkopf2021causal}.
Meta-learning and multi-task robotics improve average adaptation but typically still require target interaction for each new mix rather than algebraic reuse of singles~\cite{finn2017maml}.
System identification classically fits physical parameters; our dynamics operator is a residual closed-loop corrector under ShiftSpec, not a full inertial rebuild~\cite{ljung1999sysid}.
ManiSkill provides the controlled manipulation bed~\cite{mu2021maniskill}.
IGA and fixed freeze-$f$ recipes address module selection; our negative section shows selection alone is insufficient for the mixed-shift claim we target.

\section{LIMITATIONS}

Affine residual dynamics are an approximation under contact and long delay; under policy$\times$delay($=2$), Abelian residual compose regresses while ordered lead recovers oracle.
Vision localization is geometric (RGB-D / segmentation), not a learned perception stack; the FSM is scripted.
L1 goal localization depends on a camera-visible marker; agent-only / shoulder-only can fail under occlusion or empty views; P1 fusion removes the default workspace cam but still relies on geometric seg and a second robot camera.
Contact heuristics can disagree with true grasp; simulator success remains scoring-only; EE is proprioceptive.
On PickCube / PushCube / Peg joint8 / Peg pose7, mix$=0$ (or forced mix$=0$ for open-loop / online MP-derived replay) means we must not market residual action blending as the success driver---obs/policy composition carries the gain.
PegInsertion joint8 uses MP trajectory \emph{replay}; Peg pose7 uses online joint$\to$EE conversion with CSA on the student---neither is a MP-free reactive pose servo.
Additive generators commute by construction---delay non-commutativity is shown on the action-pipeline order.
We do not claim solved real-world arbitrary shifts, Wan/DiT-internal factors, or end-to-end learned policies.
P4 real-robot transfer is deferred pending hardware.

\paragraph{Claim audit (can / cannot).}
\textbf{Can:} compose single-factor ops for held-out mixes without mixed finetune on StackCube, PickCube (incl.\ L1/P1), PushCube, Peg joint8 replay, and Peg pose7 online EE convert; thin/deep/vision-localized BC under frozen CSA; fair adapts (mixed-singles / domain-arith) do not close the gap; ordered necessity under delay.
\textbf{Cannot:} arbitrary open-world shifts; Wan/DiT physics; mix$>0$ as the Peg/Pick win mode; MP-free reactive 7D Peg servo; real-robot P4; learned detectors as the L1 story; treating held-out mix finetune as a fair CSA baseline; pixels-to-action CNN as the P3-d win mode.

\section{CONCLUSION}

Compositional Shift Algebra reframes mixed robot shifts as operator composition rather than per-mix finetuning or module gating alone.
On StackCube, composing single-factor residual corrections extrapolates held-out mixes to oracle-level closed-loop success, extends to RGB-D vision-in-the-loop control, and exhibits ordered necessity under action delay.
On PickCube, residual CSA transfers and L1 vision control retains compositional gains without privileged cube/goal/contact in the loop, under the disclosed localization scope of Sec.~\ref{sec:threat}.
PushCube, Peg joint8/pose7, thin/deep-BC, and adapt-baseline upgrades (Sec.~\ref{sec:upgrade}) broaden task and controller claims without mixed finetuning.
IGA is reported as a negative control.
The threat model, failure modes, and main-track upgrade path above define what the present evidence supports---and what it does not.
\section*{ACKNOWLEDGMENT}
The authors thank colleagues at Harvest Praxis for discussions.

\end{document}